\documentclass[conference]{IEEEtran}
\IEEEoverridecommandlockouts
\usepackage{cite}
\usepackage[hyphens]{url}
\usepackage{amsmath,amssymb,amsfonts}
\usepackage{algorithmic}
\usepackage{graphicx}
\usepackage{subfig} 
\usepackage{textcomp}
\usepackage{booktabs}
\usepackage{multirow}
\usepackage{xcolor}
\usepackage{tabularx, booktabs, longtable, booktabs, makecell} % Tables
\usepackage{algorithmic}
\usepackage{tcolorbox}
\usepackage{float}
\usepackage{soul}
\def\BibTeX{{\rm B\kern-.05em{\sc i\kern-.025em b}\kern-.08em
    T\kern-.1667em\lower.7ex\hbox{E}\kern-.125emX}}
\newcommand\blfootnote[1]{%
  \begingroup
  \renewcommand\thefootnote{}\footnote{#1}%
  \addtocounter{footnote}{-1}%
  \endgroup
}

\begin{document}

\title{Multi-Resolution Feature Stem for Diabetic Retinopathy lesion segmentation}

\author{\IEEEauthorblockN{Indranil Dutta, Taehee Jeong}
\IEEEauthorblockA{\textit{San Jose State University} \\
indranil.dutta, taehee.jeong @sjsu.edu}
{\thanks{This work was supported in part by a Mobilint Grant awarded to San Jose State University. (Corresponding author: Taehee Jeong)}
}
}

\maketitle

\begin{abstract}
Diabetic Retinopathy (DR) is a leading cause of preventable blindness worldwide, requiring automated lesion segmentation using deep learning models for early detection and monitoring. However, DR lesions vary dramatically in size from tiny microaneurysms to large hemorrhages and exudates. This variability creates conflicting demands on the model architecture and input resolution, posing a challenge for effective design. This work investigates the impact of input resolution on different lesion types. Through systematic experimentation with multiple architectures (U-Net, UNet++, Vision Transformers, DeepLabV3+) at $512 \times 512$ and $1024 \times 1024$ resolutions, we identify a critical, counter-intuitive phenomenon where increasing input resolution has opposing effects on different lesion types.  We demonstrate that while higher resolution is essential for resolving fine-grained microaneurysms, it can unexpectedly degrade performance on larger hemorrhages. This finding challenges the common assumption that higher resolution is uniformly beneficial. To address this, we propose a novel Multi-Resolution Feature Stem, an input-level pyramid integrated with a UNet++ backbone. This architecture processes multiple scales in parallel, capturing fine-grained details without sacrificing contextual information. This work contributes crucial empirical evidence of this complex, resolution-dependent behavior and a practical, parameter-efficient architecture that successfully resolves this trade-off. 
Our code is available at \url{https://github.com/taeheej/Multi-Resolution-Feature-Stem-for-Diabetic-Retinopathy-lesion-segmentation}.
\end{abstract}


\section{Introduction}

\renewcommand{\footnoterule}{%
  \kern -3pt
  \hrule width 3in height 1pt
  \kern 2pt
}
\blfootnote{2026 International Conference on Advances in Artificial Intelligence and Machine Learning (AAIML), 20-22 March 2026, IEEE Copyright 2026}

Diabetic retinopathy (DR) is a microvascular complication of diabetes mellitus and remains one of the leading causes of preventable blindness among working-age adults worldwide \cite{teo2021global}. 589 million adults are currently living with diabetes globally. The number is projected to reach 853 million by 2050 \cite{idf2024}. 
This increase is particularly pronounced in Asian populations, where traditional diets rich in carbohydrates combined with rapid urbanization have contributed to alarming rates of type 2 diabetes. %In South and East Asian countries, the prevalence rates of diabetes have increased substantially in the past two decades, creating an unprecedented demand for retinopathy screening services. %The condition is characterized by progressive damage to the retinal blood vessels, manifesting as various types of lesions visible on fundus photographs, as illustrated in Figure~\ref{fig:fundus_examples}. 
%Early detection and precise segmentation of these lesions —including microaneurysms, hemorrhages, hard exudates, and soft exudates—will enables 
Early detection and timely monitoring of disease progression are required to potentially prevent vision loss in millions of patients \cite{cdc2020}.

\begin{figure}[htbp]
\centering
\subfloat[Healthy Retina\label{fig:dr_ex_healthy}]{%
  \includegraphics[width=0.32\linewidth]{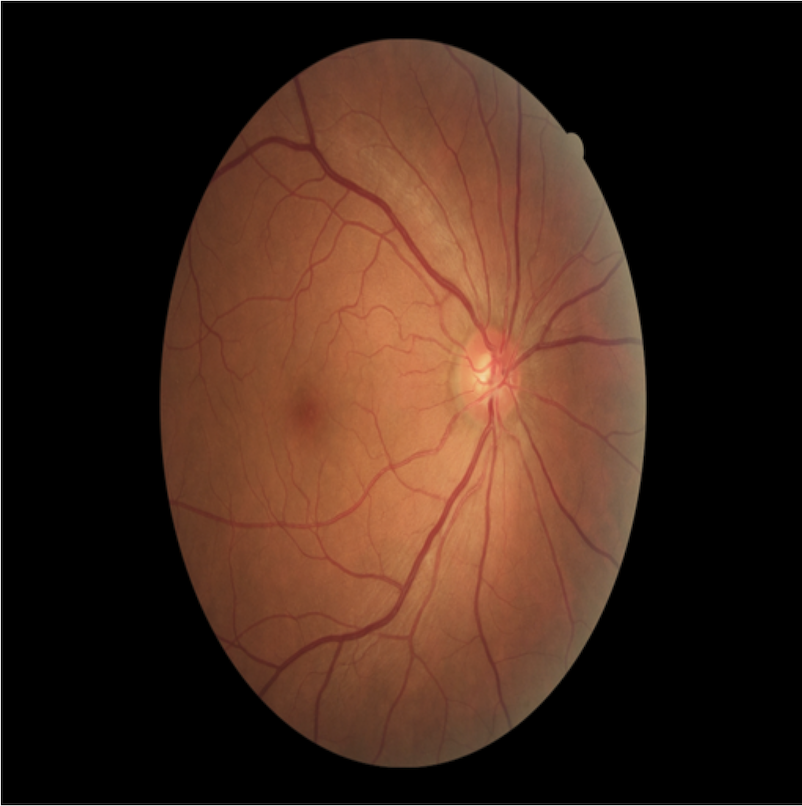}%
}\hfill%
\subfloat[Early DR with microaneurysms\label{fig:dr_ex_early}]{%
  \includegraphics[width=0.32\linewidth]{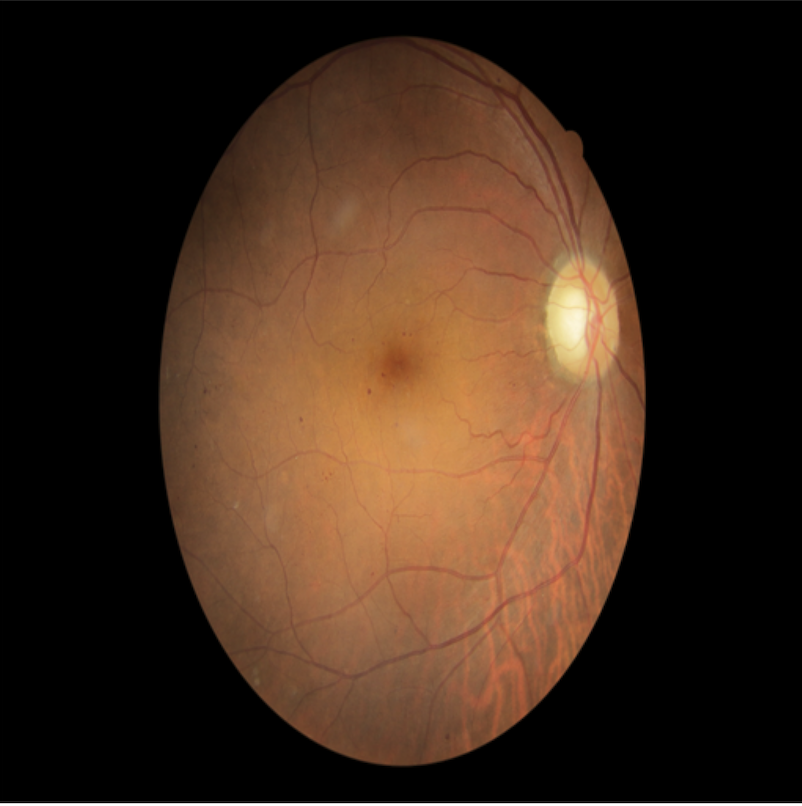}%
}\hfill%
\subfloat[Moderate DR with hemorrhages and hard exudates\label{fig:dr_ex_mod}]{%
  \includegraphics[width=0.32\linewidth]{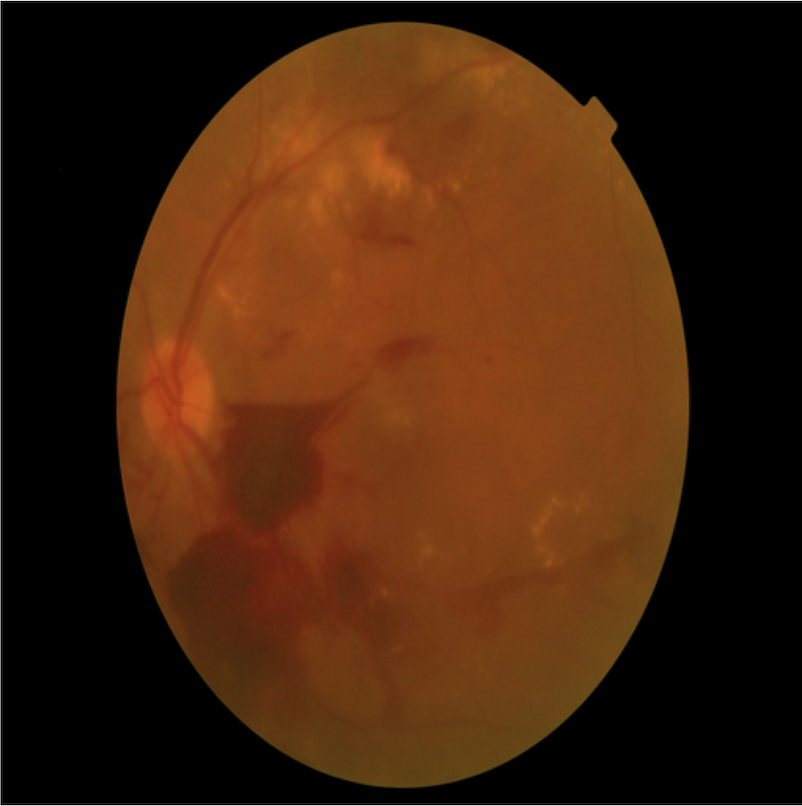}%
}\\[0.5em]
\subfloat[Severe DR with soft exudates and cotton wool spots\label{fig:dr_ex_sev}]{%
  \includegraphics[width=0.32\linewidth]{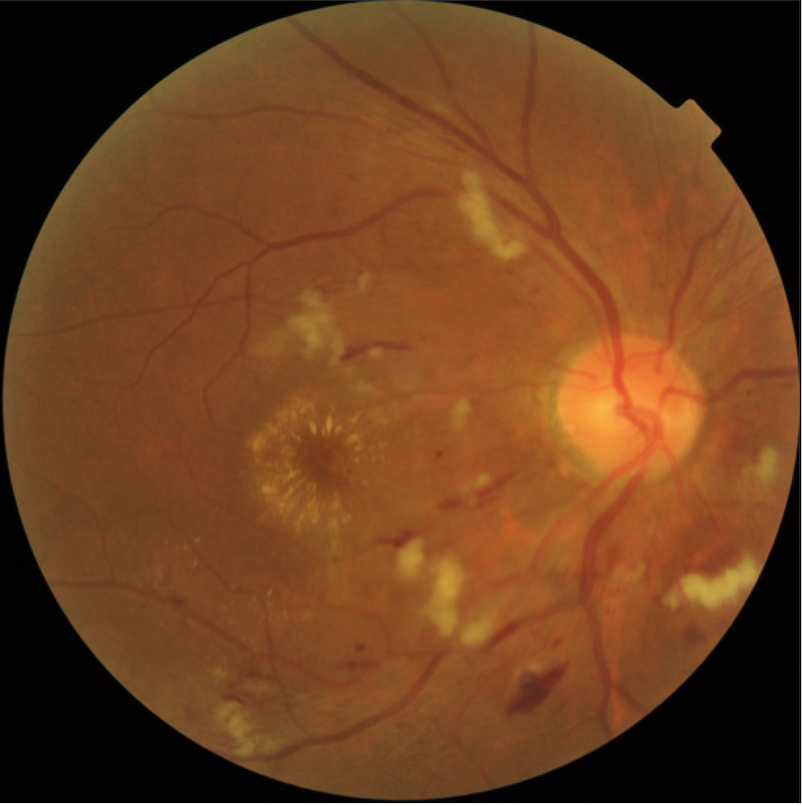}%
}\hfill%
\subfloat[Proliferative DR\label{fig:dr_ex_prol}]{%
  \includegraphics[width=0.32\linewidth]{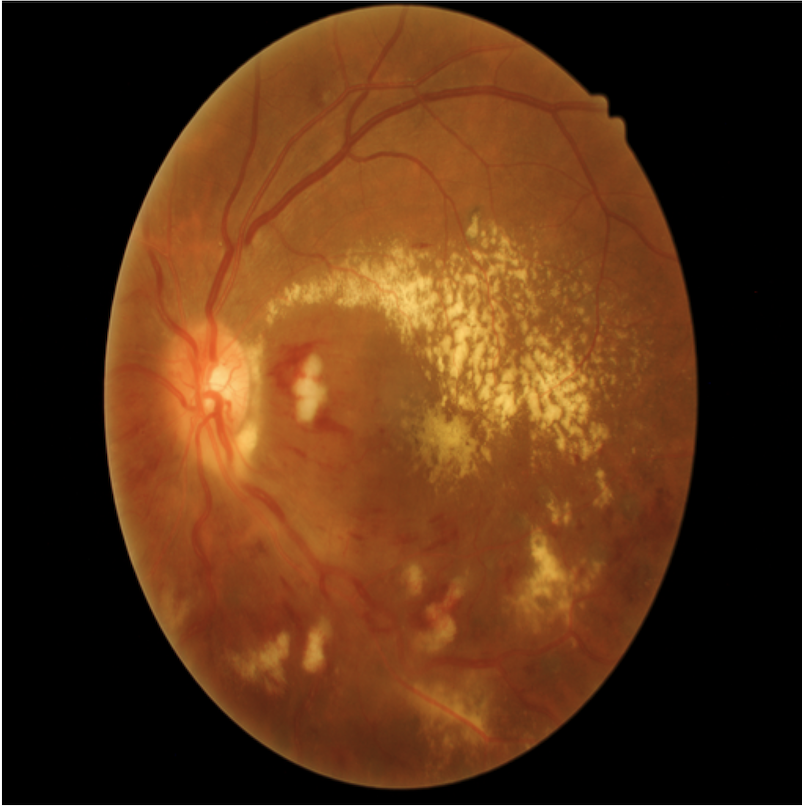}%
}\hfill%
\subfloat[Varying lesion types and severity\label{fig:dr_ex_vary}]{%
  \includegraphics[width=0.32\linewidth]{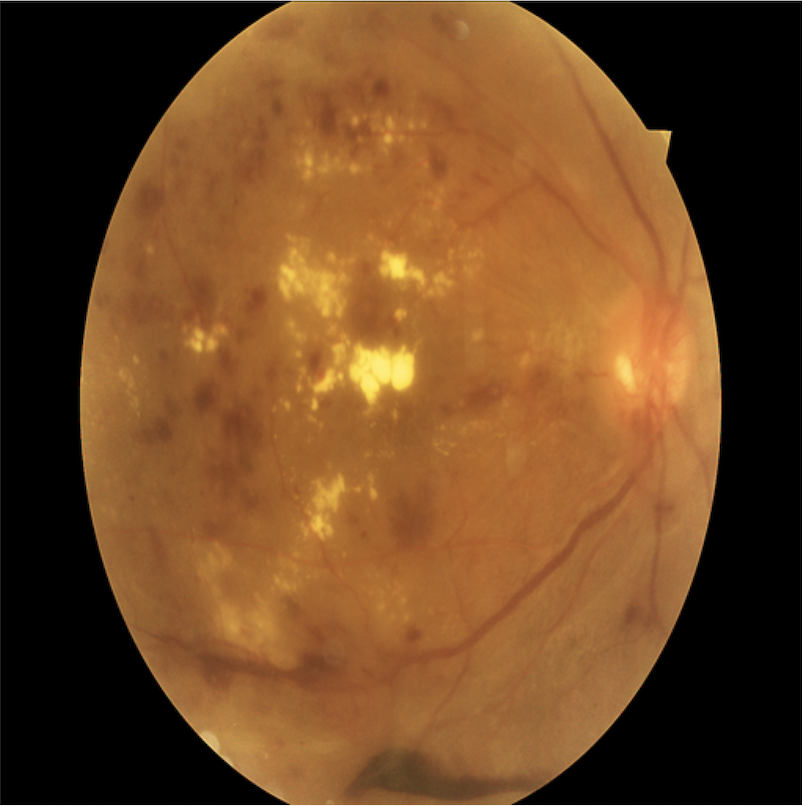}%
}
\caption{Fundus photographs illustrating various DR lesions}
\label{fig:fundus_examples}
\end{figure}

% \begin{figure}[htbp]
%   \centering
%   \includegraphics[width=0.45\textwidth]{images/dr_severity_examples_labeled.png}
%   \caption{Fundus photographs illustrating various DR lesions. (a) Healthy retina. (b) Early DR with microaneurysms. (c) Moderate DR with hemorrhages and hard exudates. (d) Severe DR with soft exudates and cotton wool spots. (e) Proliferative DR. (f) Another example of varying lesion types and severity. }
%   \label{fig:fundus_examples}
% \end{figure}

Current clinical practice for the screening of diabetic retinopathy involves manual examination and annotation of fundus images by trained ophthalmologists. This process is time-consuming, expensive, and creates significant bottlenecks in patient care. Patients often experience delays of two weeks or longer between image acquisition and receiving diagnostic results \cite{brennan2024addressing}. This delays might result in missed opportunities for early intervention during critical disease stages \cite{scanlon2014delay}. Furthermore, manual annotation is subject to inter-observer variability, with studies showing disagreement rates of about 15\% among expert retina specialists \cite{srinivasan2023inter}. %, and inter-observer reliability kappa scores ranging from 0.22 to 0.91 in different grading contexts \cite{teoh2023variability}. 
%The shortage of trained specialists, particularly in underserved and rural areas, exacerbates these challenges \cite{yau2012global}. 

Automated lesion segmentation using deep learning models offers the potential for scalable, consistent, and cost-effective screening that could provide near-immediate diagnostic feedback. However, %the development of robust automated segmentation systems faces significant technical challenges that have limited their clinical adoption, including the 
significant technical challenges remain for accurately detecting and segmenting diverse types of lesions with varying sizes and characteristics. 
%Deep learning approaches to diabetic retinopathy lesion segmentation face substantial challenges arising from the diverse physical characteristics of different types of lesions. 
Microaneurysms (MA), representing early-stage diabetic retinopathy, appear as small circular red dots often measuring between 20 and 125 micrometers in diameter \cite{dubow2014classification}, making them among the smallest pathological features requiring detection in fundus images. In contrast, hemorrhages (HE) vary considerably in size and shape, while exudates (EX) present as bright yellow-white deposits with irregular boundaries \cite{wilkinson2003proposed}.

The varying characteristics of lesions pose a challenge for deep learning model architecture and input resolution, as they create competing demands. Traditional medical image segmentation approaches typically process images at a fixed input resolution, but this method has a significant limitation since lesion sizes can vary significantly depending on the type of lesion.

To address the complex resolution sensitivities, we proposes a novel Multi-Resolution Feature Stem. Building upon UNet++ \cite{Zhou2018} backbone, we develop a modular framework that simultaneously leverages features from multiple input resolutions. The proposed approach constructs an input pyramid at multiple scales ($1024 \times 1024$, $512 \times 512$, $256 \times 256$), processes each resolution through shared Convolutional layers, fuses the resulting multi-scale features into a single rich tensor, and integrates this tensor to the first layer of the UNet++ encoder. This design enables the model to benefit from information at multiple scales simultaneously, addressing the empirical finding that different lesion types have competing resolution requirements.

%To address the complex resolution sensitivities identified through our empirical analysis, we proposes a novel Multi-Resolution Feature Stem. Building upon UNet++\cite{Zhou2018}, which demonstrated strong baseline performance, we develop a modular framework that simultaneously leverages features from multiple input resolutions. The proposed approach constructs an input pyramid at multiple scales ($1024 \times 1024$, $512 \times 512$, $256 \times 256$), processes each resolution through shared Convolutional layers, fuses the resulting multi-scale features into a single rich tensor, and integrates this tensor by surgically modifying the first layer of the UNet++ encoder to accept the high-dimensional input. This design enables the model to benefit from information at multiple scales simultaneously, addressing the empirical finding that different lesion types have competing resolution requirements.

\section{Related Work}
%Recent developments in deep learning for semantic segmentation have revolutionized lesion detection in fundus images, offering significant improvements over traditional handcrafted feature approaches. However, single-resolution, single-pathway architectures may struggle with the heterogeneous characteristics of DR lesions.  Thus, several researchers have proposed multi-input and dual-path approaches that process images at different scales or through parallel pathways. %These methods represent an important step toward acknowledging that uniform processing strategies may be insufficient for multi-class lesion segmentation.

The application of deep learning to semantic segmentation has significantly enhanced lesion detection in fundus images, surpassing the limitations of traditional handcrafted feature-based approaches. Nevertheless, traditional single-resolution, single-pathway architectures \cite{aruna2025}  often struggle to accurately capture the diverse characteristics of DR lesions. To address this challenge, researchers have explored alternative approaches, including multi-input and dual-path architectures that process images at multiple scales or through parallel pathways, allowing for more effective detection of lesions with varied characteristics.

CARNet \cite{guo2022carnet} employs a dual-input encoder-decoder structure using both whole fundus images and cropped patches to achieve multi-lesion segmentation. The dual-input design enables the model to capture both global context from full images and fine-grained details from local patches. Attention Refinement Modules (ARMs) fuse global and local features, enhancing segmentation accuracy by combining information across spatial scales. %It demonstrated robustness and accuracy across multiple datasets, suggesting that multi-scale input processing offers advantages over single-scale approaches.

%DARNet (Dual-input Attentive RefineNet) \cite{guo2022multiple}  similarly utilizes a dual-input encoder-decoder structure to address the challenges posed by complex lesion structures and varying lesion sizes. The dual-input approach processes information at different scales in parallel, with Attention Refinement Modules further enhancing the segmentation process by selectively emphasizing relevant features from each pathway.

Many-to-Many Reassembly of Features (M2MRF) \cite{liu2023automated} address challenges in segmenting tiny retinal lesions. M2MRF introduces a many-to-many feature reassembly mechanism that preserves subtle activations and captures long-term dependencies, preventing the loss of information about small lesions during feature propagation. %When applied to HRNetV2, 
%M2MRF achieves improved results and generalization abilities compared to existing methods, demonstrating that careful preservation of multi-scale information is critical for detecting lesions across size ranges.

PMCNet \cite{he2022progressive} integrates progressive feature fusion and dynamic attention blocks to address multi-scale segmentation challenges. The progressive feature fusion block facilitates feature learning by aggregating fine-grained details from early layers with high-level semantics from deeper layers. The dynamic attention block mitigates feature inconsistencies that arise when combining multi-scale representations. %This method demonstrates significant performance gains in multi-class lesion segmentation, highlighting the value of carefully designed feature fusion strategies.

While these multi-input and dual-path approaches recognize the need for multi-scale processing, they apply uniform fusion strategies across all lesion types. Both input streams or pathways process MA, HE, and EX identically, with the same multi-scale features combined in the same manner regardless of lesion type. Furthermore, existing multi-scale approaches typically operate at the feature level within the network architecture, rather than systematically investigating how input resolution itself affects different lesion types. %Whether different lesions exhibit differential sensitivity to input resolution scaling and what architectural strategies might address such differential behavior remains unexplored.

\section{Methodology}

\subsection{Dataset}
This work utilizes the Diabetic Retinopathy Dataset (DDR) \cite{Li2019}, a large and widely recognized collection of fundus images. The DDR dataset has become a standard benchmark in the field, not only for DR grading but also for lesion segmentation, owing to its comprehensive annotations and high image quality. The specific subset employed in this work is the lesion segmentation task, which provides %ground truth masks for the four most clinically significant DR-related pathologies. 
%The dataset includes 
pixel-level annotations for four lesion types, each presenting unique morphological characteristics and challenges for automated segmentation algorithms. %A brief description of each class is as follows:

\begin{itemize}  % need references
  \item \textbf{Exudates (EX):} These are lipid residues that leak from damaged retinal blood vessels. They typically appear as bright, yellowish-white patches with relatively well-defined borders. Their size can vary significantly, from small dots to large confluent areas %(Figure \ref{fig:dataset_ex_train}).
  \item \textbf{Hemorrhages (HE):} These are areas of bleeding from damaged vessels into the retinal layers. They appear as red spots and can manifest in various shapes and sizes, from small "dot-and-blot" hemorrhages to larger, flame-shaped bleeds. Their appearance can be challenging to distinguish from the underlying retinal vasculature %(Figure \ref{fig:dataset_he_train}).
  \item \textbf{Microaneurysms (MA):} These are tiny, localized balloon-like swellings of retinal capillaries and are often the earliest clinical sign of diabetic retinopathy. They appear as small, circular, reddish dots, typically ranging from 20 to 125 micrometers in diameter. Their minuscule size makes them exceptionally challenging to detect %(Figure \ref{fig:dataset_ma_train}).
  \item \textbf{Soft Exudates (SE):} Also known as cotton wool spots, these are nerve fiber layer infarctions that appear as fluffy, white patches with indistinct borders. They are generally larger than microaneurysms but can be confused with hard exudates or optic disc reflections %(Figure \ref{fig:dataset_se_train}).
\end{itemize}

The DDR dataset is pre-partitioned into Training Set (383 images), Validation Set (149 images), and Test Set (225 images).
%The DDR dataset is pre-partitioned into three distinct sets, which were used as follows:
%\begin{itemize}
%\item The Training Set (383 images), used exclusively for model training. Data augmentation
%was applied solely to this partition.
%\item The Validation Set (149 images), used to evaluate performance after each epoch for
%hyperparameter tuning and model selection. The best-performing checkpoint on this set
%was saved for final evaluation.
%\item The Test Set (225 images), held out during all training and selection phases. It %provides
%the final, unbiased assessment of the selected model’s generalization capability.
%\end{itemize}

\subsection{Data Preprocessing}
The fundus photographs were provided in the standard JPG format as 3-channel RGB images. The corresponding ground truth annotations were provided as separate, single-channel binary TIF masks for each of the four lesion types. These individual binary masks were stacked, resulting in a 4-channel ground truth tensor of the same height and width as the input image. Each channel in this tensor corresponds to one lesion type (EX, HE, MA, SE), and the value at each pixel location is either 1 (lesion present) or 0 (lesion absent).

%To create a unified segmentation target for the multi-class models, these individual binary masks were loaded and stacked along a new channel dimension, resulting in a 4-channel ground truth tensor of the same height and width as the input image. Each channel in this tensor corresponds to one lesion class (EX, HE, MA, SE), and the value at each pixel location is either 1 (lesion present) or 0 (lesion absent).

%Our hypothesis is that input resolution has a differential impact on the segmentation performance for lesions of varying scales. To investigate this, we standardized all images to two target resolutions: $1024 \times 1024$ and $512 \times 512$. The choice of interpolation algorithm for this resampling process is critical, as an incorrect choice can introduce artifacts or destroy information.
To investigate the impact of input resolution on segmentation performance for different lesion types, we resized all images to two specific resolutions: $1024 \times 1024$ and $512 \times 512$. Since the original images had variable aspect ratios, we preserved their aspect ratios during the resizing process and then applied a center crop to obtain the desired square dimensions.

% The choice of interpolation algorithm for this resampling process is critical, as an incorrect choice can introduce artifacts or destroy information.
% \begin{itemize}
%     \item For the RGB images, Bilinear Interpolation was used. This method calculates the value of a new pixel based on a weighted average of the four nearest pixels in the original image. It provides a smooth transition in pixel intensities and is a standard choice for downsampling natural images without introducing the blocky artifacts associated with simpler methods.
%     \item For the 4-channel ground truth masks, Nearest-Neighbor Interpolation was employed. This method assigns the value of the nearest pixel in the original image to the new pixel, without any averaging. This choice is crucial for segmentation masks, as it guarantees that no new pixel values are created. Using an averaging method like bilinear interpolation would introduce intermediate floating-point values (e.g., 0.5) between the binary 0 and 1 labels, corrupting the discrete class information of the ground truth.
% \end{itemize}

The choice of interpolation algorithm for this resampling process is critical, as an incorrect choice can introduce artifacts or destroy information. For the RGB images, bilinear interpolation was used to provide smooth pixel intensity transitions without blocky artifacts. For the 4-channel ground truth masks, nearest-neighbor interpolation was employed to preserve the discrete binary labels, as averaging methods would introduce intermediate floating-point values that corrupt the class information.

Given the limited training set size (383 images), data augmentation was critical to prevent overfitting. We applied a suite of transformations, detailed in Table \ref{tab:augmentations}, to artificially expand the dataset. These included geometric transformations (flips, rotation, scaling) to teach orientation invariance, photometric transformations (color jitter) to handle illumination variance, and structural transformations (elastic deformation, blur) to simulate tissue warping and focus imperfections.

\begin{table}[ht]
\centering
\caption{Data Augmentation Techniques and Parameters Applied to the Training Set}
\label{tab:augmentations} 
\begin{tabular}{p{2cm} p{4cm} p{1cm}}
\toprule
\textbf{Augmentation Technique} & \textbf{Parameters} & \textbf{Probability} \\
\midrule
Horizontal Flip & None & 0.5 \\
Vertical Flip & None & 0.5 \\
Random Rotation & Rotation range: $\pm 15^{\circ}$ & 0.5 \\
Random Scale & Scale limit: 0.9 -- 1.1 & 0.3 \\
  Color Jitter & Brightness=0.2, Contrast=0.2, Sat.=0.1, Hue=0.05 & 0.3 \\
Elastic Deformation & alpha=120, sigma=120 & 0.2 \\
Gaussian Blur & kernel\_size=3, sigma=(0.1, 2.0) & 0.2 \\
\bottomrule
\end{tabular}
\end{table}

%\begin{table}[ht]
%\centering
%\caption{Data Augmentation Techniques and Parameters Applied to the Training Set.}
%\label{tab:augmentations}
%  \begin{tabular}{l >{\RaggedRight}p{5.5cm} c}
%  \toprule
%  \textbf{Augmentation Technique} & \textbf{Parameters} & \textbf{Probability} \\
%  \midrule
%Horizontal Flip & None & 0.5 \\
%Vertical Flip & None & 0.5 \\
%Random Rotation & Rotation range: $\pm 15^{\circ}$ & 0.5 \\
%Random Scale & Scale limit: 0.9 -- 1.1 & 0.3 \\
%  Color Jitter & Brightness=0.2, Contrast=0.2, Sat.=0.1, Hue=0.05 & 0.3 \\
%Elastic Deformation & alpha=120, sigma=120 & 0.2 \\
%Gaussian Blur & kernel\_size=3, sigma=(0.1, 2.0) & 0.2 \\
%  \bottomrule
%\end{tabular}
%\end{table}

\begin{figure*}[t]
  \centering
  \includegraphics[width=0.65\textwidth]{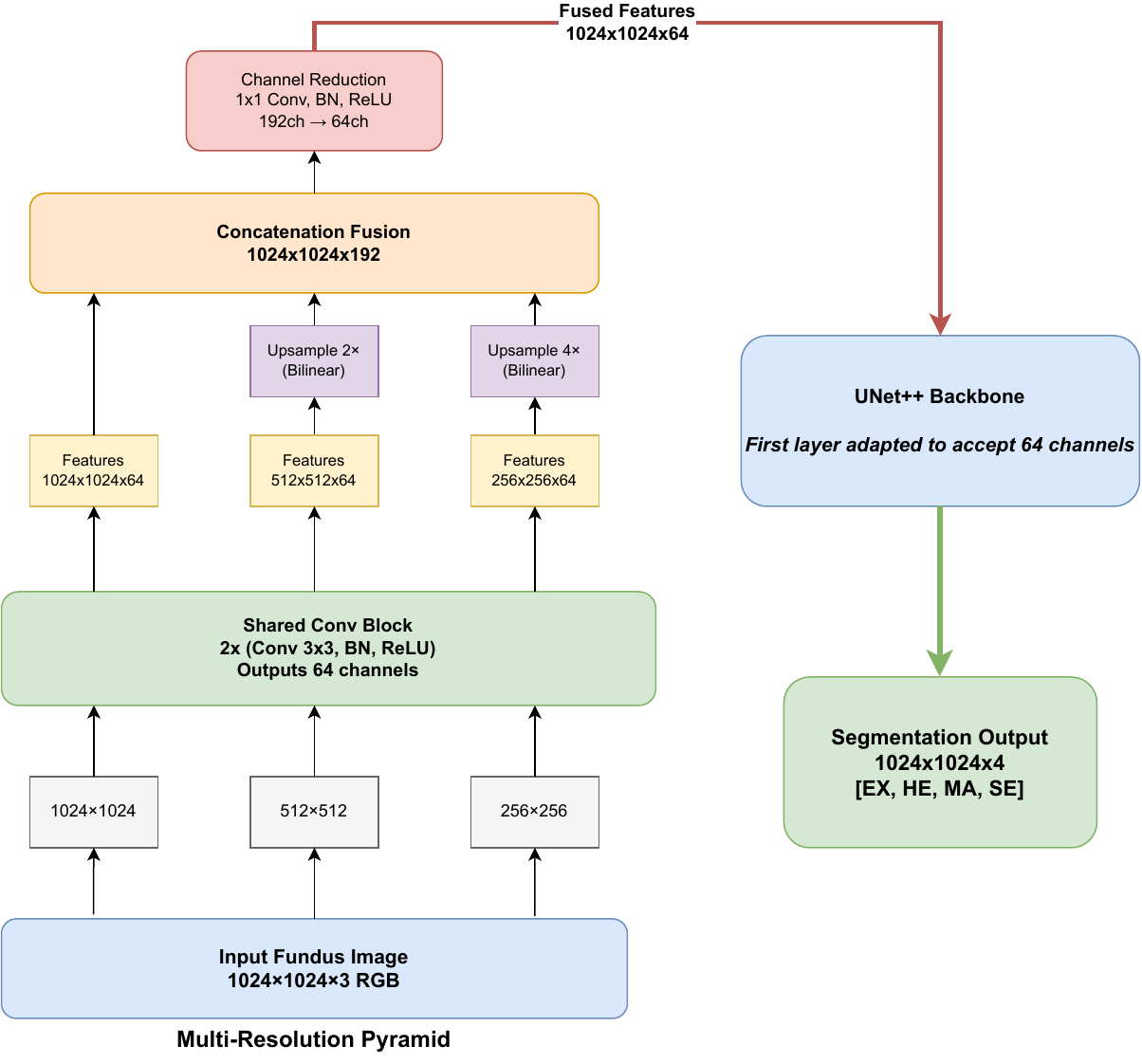}
  \caption{Multi-Resolution Feature Stem architecture}
  \label{fig:proposed_arch}
\end{figure*}

\subsection{Model Architectures}

We propose a novel Multi-Resolution Feature Stem that is prepended to a UNet++ backbone. A detailed diagram of this architecture is presented in Figure~\ref{fig:proposed_arch}. The guiding principle is to process the input at multiple resolutions in parallel, forcing the model to learn scale-invariant features. These features are then fused into a single, rich representation that is simultaneously aware of fine-grained detail and broad semantic context \textit{before} the main encoder begins its work.

%\begin{figure}[htbp]
%  \centering

%  \includegraphics[width=\textwidth]{images/architecture_diagram.pdf}
  
%  \caption[The proposed Multi-Resolution Feature Stem architecture.]{The proposed Multi-Resolution Feature Stem architecture. The input image is processed at three scales in parallel using a \textbf{shared-weight} convolutional block. The resulting feature maps are upsampled, concatenated, and fused by a 1x1 convolution. This 64-channel, multi-scale tensor is then fed into a \textbf{modified} UNet++ backbone whose first layer is adapted to accept the 64-channel input.}
%  \label{fig:proposed_arch}
%\end{figure}

The Multi-Resolution Feature Stem replaces the input stage of the UNet++ backbone. Its design consists of three key stages:

\begin{itemize}
    \item Parallel Input Pyramid: From a single high-resolution $1024 \times 1024$ input image, a three-level image pyramid is generated via bilinear down-sampling, resulting in parallel inputs at $1024 \times 1024$, $512 \times 512$, and $256 \times 256$.

    \item Shared-Weight Feature Extraction: Each of the three parallel streams is processed by an identical, shared-weight Convolutional block (consisting of two $3 \times 3$ convolutions). This parameter-efficient design, which was a key of our work, encourages the network to learn a single set of universal, scale-invariant features. Each stream emerges from this shared block as a 64-channel feature map.

    \item Pre-Backbone Fusion and Adaptation: The feature maps from the $512 \times 512$ and $256 \times 256$ streams are up-sampled (using bilinear interpolation) to match the $1024 \times 1024$ resolution. All three 64-channel feature maps are then concatenated along the channel dimension, resulting in a single, high-dimensional $192$-channel tensor. A $1 \times 1$ convolution then fuses and reduces this tensor to a compact $64$-channel representation.

    \item Backbone Integration: To seamlessly integrate this rich, multi-scale tensor, the first Convolutional layer of the UNet++ backbone (originally designed to accept a 3-channel RGB image) is replaced with a new Convolutional layer built to accept the $64$-channel fused input from our stem, allowing the entire encoder to proceed with a feature representation that is multi-scale from the very first layer.
\end{itemize}

% \subsection{Optimizer and Learning rate}
% For all experiments, we employed the AdamW optimizer \cite{loshchilov2017decoupled}  and set the initial learning rate to $2 \times 10^{-4}$.  The optimizer was configured with a weight decay rate of $0.01$ and the standard momentum parameters of $\beta_1 = 0.9$ and $\beta_2 = 0.999$. For the initial training phase, the Cosine Annealing Learning Rate Scheduler was used. This scheduler smoothly decays the learning rate from an initial maximum to a minimum value ($\eta_{min}$) of $1 \times 10^{-6}$ over a period ($T_{max}$) of 50 epochs. For the subsequent fine-tuning phase, we switched to the OneCycle Learning Rate Scheduler \cite{smith2019onecycle}. The OneCycle policy, which gradually increases the learning rate from a low value to a maximum and then decreases it, has been shown to act as a powerful regularization method and can lead to faster convergence and better final performance. This switch allowed the model to escape any suboptimal local minima found during the initial cosine decay and more effectively explore the loss landscape for a better solution. 

% Initially, all models were trained for a fixed duration of 50 epochs. After the initial 50-epoch run, training was resumed using the OneCycleLR scheduler, allowing the models to continue learning and push performance metrics higher until the validation mAP demonstrated no further improvement.

\subsection{Loss Function}

 The Focal Loss  \cite{lin2017focal} was introduced as an enhancement to the standard Cross-Entropy Loss to address class imbalance in object detection, and its properties are highly applicable to segmentation. The Focal Loss is defined as:
 \begin{equation}
 \mathcal{L}_{\text{Focal}}(p_t) = - \alpha_t (1 - p_t)^\gamma \log(p_t)
 \label{eq:focal_loss}
 \end{equation}
 In our experiments, we used the common hyperparameter settings of a focusing parameter $\gamma = 2.0$ and a weighting factor $\alpha = 0.75$ for the positive classes.

 While Focal Loss addresses the class imbalance from a classification perspective, it does not directly optimize for the spatial overlap that is critical for segmentation quality. For this, we incorporate the Dice Loss \cite{dice1, dice2}, which is derived from the Sørensen–Dice coefficient, a metric used to gauge the similarity of two samples. The Dice Coefficient is defined as:
 \begin{equation}
 \text{DSC} = \frac{2 \times |A \cap B|}{|A| + |B|}
 \label{eq:dice_coefficient}
 \end{equation}
 where $A$ and $B$ are the predicted and ground truth segmentation masks, respectively. In practice, a small smoothing constant $\epsilon = 1$ is added to the numerator and denominator to ensure numerical stability, yielding the Dice Loss:
 \begin{equation}
 \mathcal{L}_{\text{Dice}} = 1 - \frac{2 \times |A \cap B| + \epsilon}{|A| + |B| + \epsilon}
 \label{eq:dice_loss}
 \end{equation}

 We implemented a weighted sum of the Focal Loss and the Dice Loss, designed to provide a balanced optimization objective. The formulation is as follows:
\begin{equation}
\mathcal{L}_{\text{total}} = 0.6 \times \mathcal{L}_{\text{Focal}} + 0.4 \times \mathcal{L}_{\text{Dice}}
\label{eq:combined_loss}
\end{equation}
The 60/40 weighting was found to provide the best empirical performance during preliminary experiments.

 By minimizing this value, the model is directly encouraged to maximize the spatial overlap between its prediction and the ground truth. This makes it inherently robust to class imbalance, as it only considers the foreground pixels of the prediction and the target. By combining these complementary objectives, our model is trained to simultaneously address class imbalance through focused attention on hard examples while directly optimizing for precise boundary delineation and spatial overlap.

\subsection{Implementation Details}

The majority of experiments were run on an Apple MacBook Pro with an M4 Max processor, featuring a 40-core GPU and 36GB of unified memory. Training was accelerated using Apple's Metal Performance Shaders (MPS) backend. 
For more computationally intensive models, a Google Colab Pro instance with an NVIDIA A100 GPU was used. This provided 40GB of dedicated VRAM and Tensor Core acceleration. Training on this instance utilized CUDA 12.x and Automatic Mixed Precision (AMP) for a significant speedup.

\section{Experimental Results}
The initial phase of our evaluation focused on establishing performance benchmarks using three established semantic segmentation models: DeepLab-v3+\cite{chen2018encoder}, U-Net\cite{ronneberger2015unet}, and U-Net++. Each model was trained and evaluated on the validation set, and their performance was quantified using mean Average Precision (mAP), mean Intersection over Union (mIoU), and per-class Average Precision (AP) for four distinct classes: Exudates (EX), Hemorrhages (HE), Microaneurysms (MA), and Soft Exudates (SE).

\begin{figure}[ht]
\centering
\subfloat[Input\label{fig:seg_ex_1}]{%
  \includegraphics[width=0.32\linewidth]{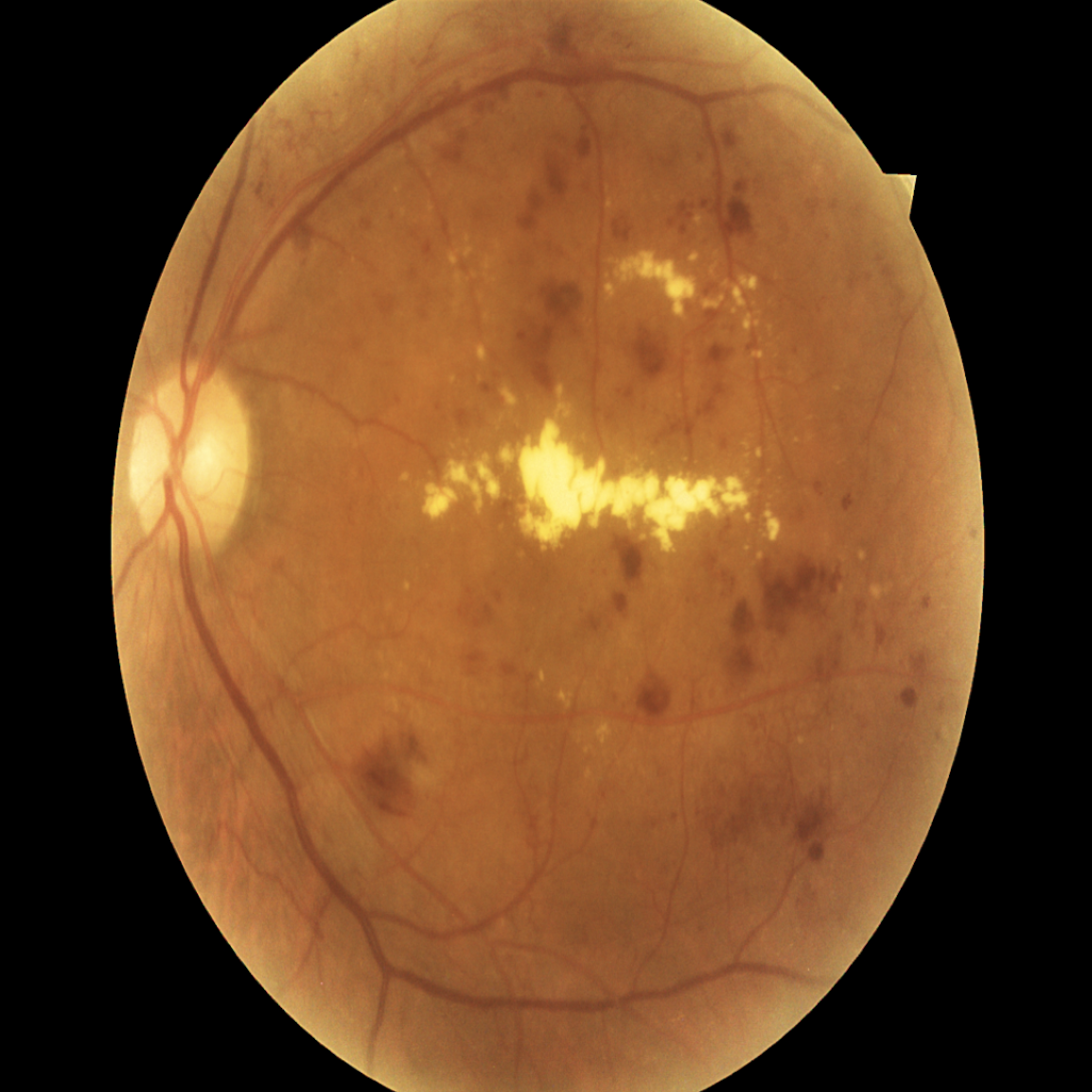}%
}\hfill%
\subfloat[Ground Truth\label{fig:seg_ex_2}]{%
  \includegraphics[width=0.32\linewidth]{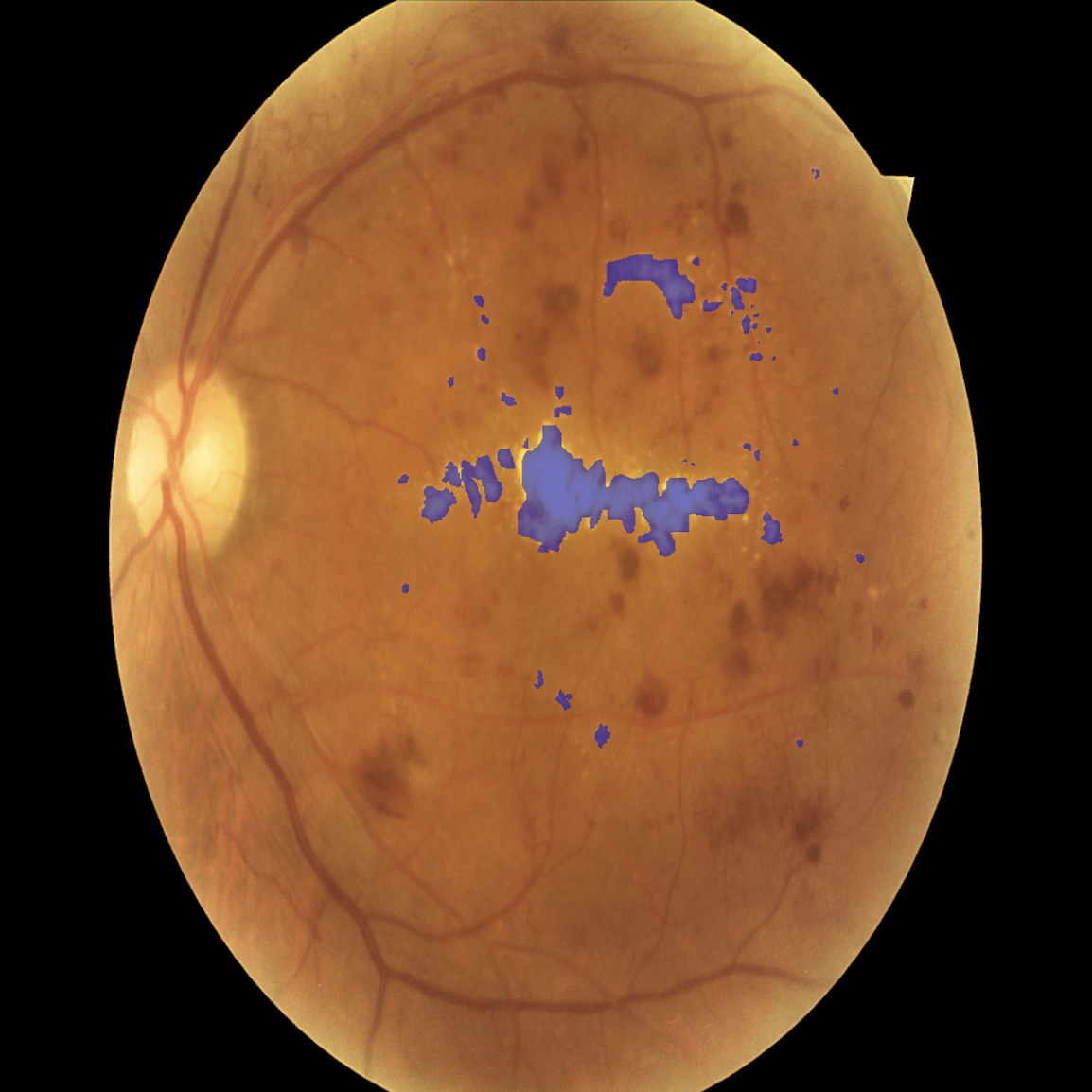}%
}\hfill%
\subfloat[Prediction\label{fig:seg_ex_3}]{%
  \includegraphics[width=0.32\linewidth]{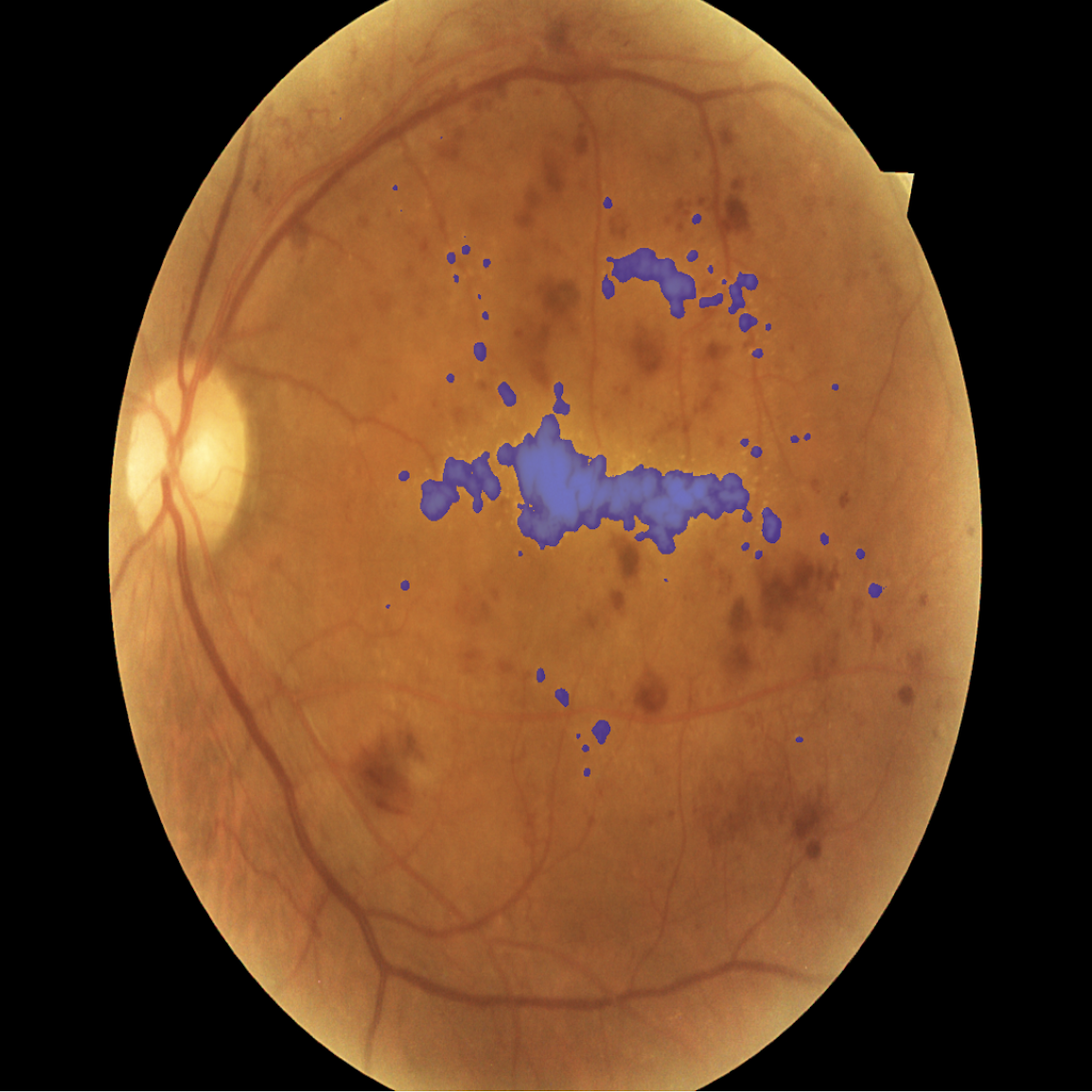}%
}
\caption{Segmentation example for Exudates}
\label{fig:ex_qual}
\end{figure}

% \begin{figure}[ht]
% \centerline{\includegraphics[width=0.95\linewidth]{images/007-3892-200_EX_comparison.png}}
% \caption{Segmentation example for Exudates: (a) Input image, (b) Ground Truth, and (c) Prediction}
% \label{fig:ex_qual}
% \end{figure}

\begin{figure}[ht]
\centering
\subfloat[Input\label{fig:seg_he_1}]{%
  \includegraphics[width=0.32\linewidth]{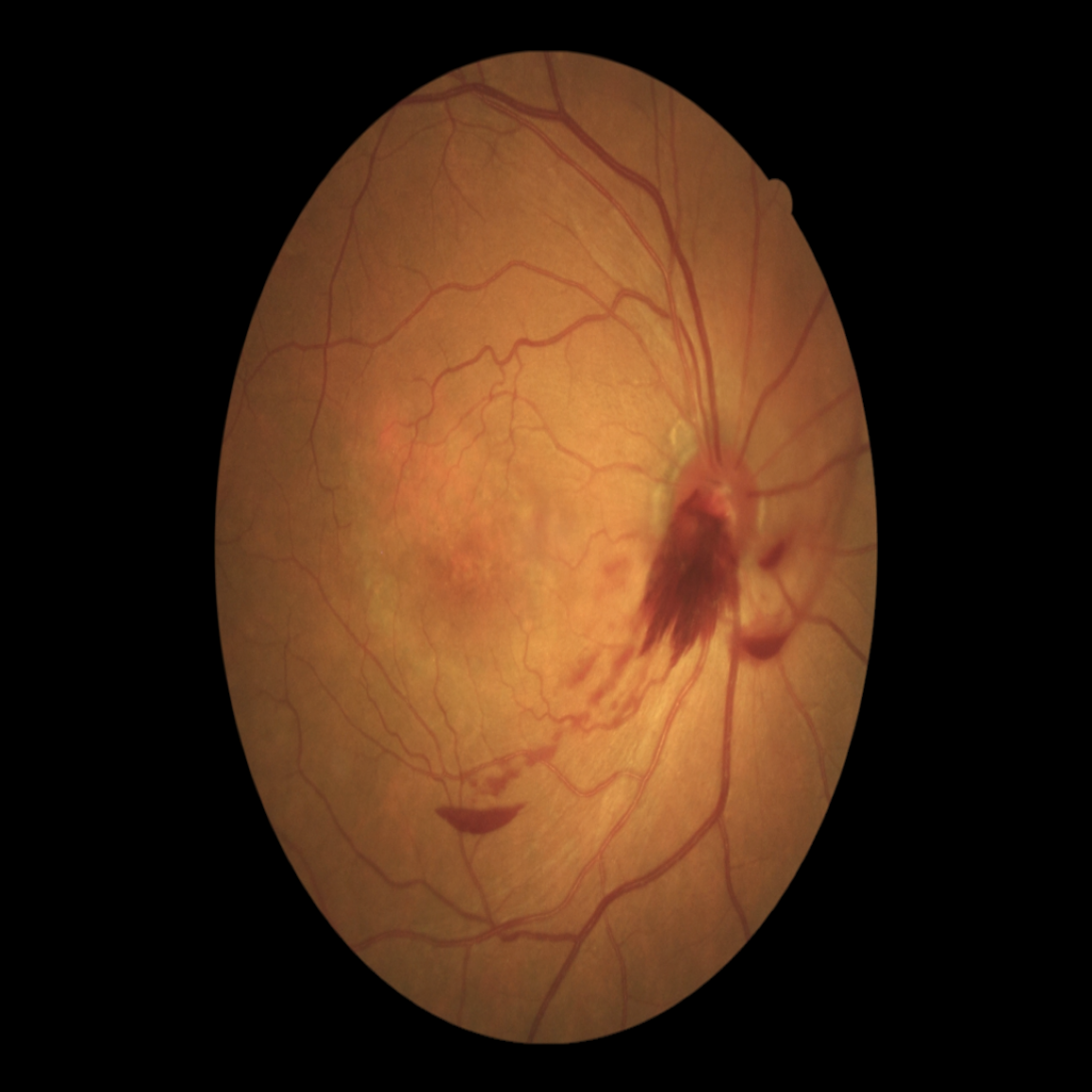}%
}\hfill%
\subfloat[Ground Truth\label{fig:seg_he_2}]{%
  \includegraphics[width=0.32\linewidth]{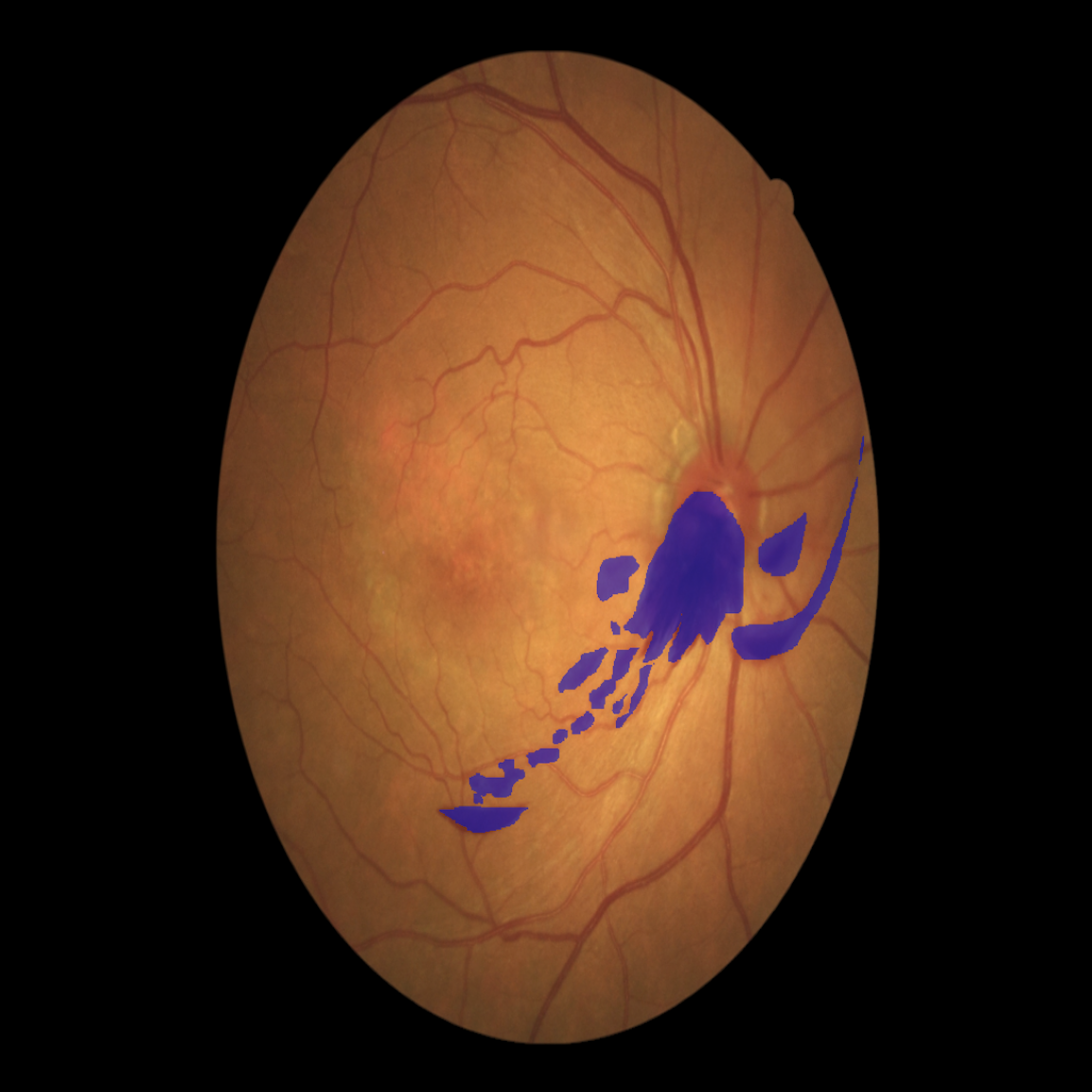}%
}\hfill%
\subfloat[Prediction\label{fig:seg_he_3}]{%
  \includegraphics[width=0.32\linewidth]{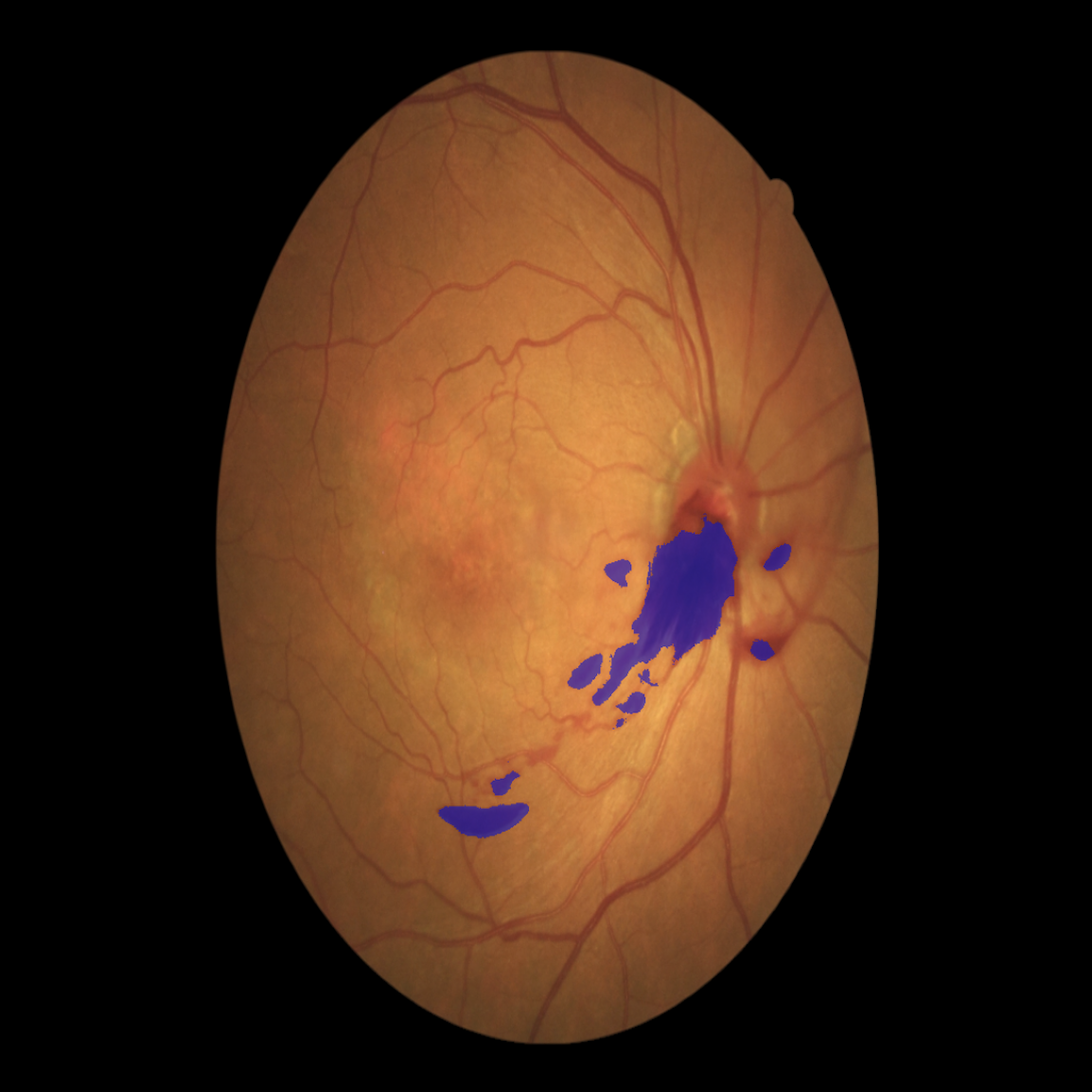}%
}
\caption{Segmentation example for Hemorrhages}
\label{fig:he_qual}
\end{figure}

% \begin{figure}[ht]
% \centerline{\includegraphics[width=0.95\linewidth]{images/007-6811-400_HE_comparison.png}}
% \caption{Segmentation example for Hemorrhages: (a) Input image, (b) Ground Truth, and (c) Prediction}
% \label{fig:he_qual}
% \end{figure}

The baseline models' performance on Validation dataset is summarized in Table~\ref{tab:baseline_performance}. Per-class metrics are reported using Average Precision (AP). The highest value in each column is highlighted in bold. 
The U-Net variants, including U-Net and U-Net++, consistently outperformed the DeepLab-v3+ architecture across all primary evaluation metrics. In terms of pixel-level segmentation accuracy, U-Net++ achieved the highest mIoU score of 0.3513, indicating the best overall performance. Meanwhile, U-Net excelled in object detection-style evaluation, recording the highest mAP score of 0.5378. In contrast, DeepLab-v3+ lagged behind, yielding the lowest scores for both mIoU (0.2098) and mAP (0.3848).

% \begin{figure}[ht]
% \centerline{\includegraphics[width=0.95\linewidth]{images/20170506085734414_MA_comparison.png}}
% \caption{Segmentation example for Microaneurysms: (a) Input image, (b) Ground Truth, and (c) Prediction}
% \label{fig:ma_qual}
% \end{figure}

% \begin{figure}[ht]
% \centerline{\includegraphics[width=0.95\linewidth]{images/007-6524-400_SE_comparison.png}}
% \caption{Segmentation example for Soft Exudates: (a) Input image, (b) Ground Truth, and (c) Prediction}
% \label{fig:se_qual}
% \end{figure}

The per-class evaluation revealed distinct strengths between the two U-Net architectures. U-Net demonstrated the best results for the Exudates (AP = 0.6121) and Hemorrhages (AP = 0.5922) classes. Conversely, U-Net++ showed superior performance in segmenting the Microaneurysms (AP = 0.3660) and Soft Exudates (AP = 0.6821) classes. This distribution of per-class strengths highlights that no single baseline model was dominant across all categories. DeepLab-v3+ did not outperform the other baselines in any individual class.

% LaTeX Table for Baseline Model Performance
\begin{table}[htbp]
\centering
\caption{Performance of Baseline Segmentation Models}
\label{tab:baseline_performance}
\resizebox{0.45\textwidth}{!}{
\begin{tabular} {lcccccc}
\hline
\textbf{Model Name} & \textbf{mAP} & \textbf{mIoU} & \textbf{AP\_EX} & \textbf{AP\_HE} & \textbf{AP\_MA} & \textbf{AP\_SE} \\ \hline
DeepLab-v3+ & 0.3848 & 0.2098 & 0.4245 & 0.4270 & 0.1976 & 0.4902 \\
U-Net       & \textbf{0.5378} & 0.3136 & \textbf{0.6121} & \textbf{0.5922} & 0.3318 & 0.6150 \\
U-Net++     & 0.5220 & \textbf{0.3513} & 0.5880 & 0.4520 & \textbf{0.3660} & \textbf{0.6821} \\ \hline
\end{tabular}
}
\end{table}

To investigate the sensitivity of the baseline models to input image size, we conducted experiments at two distinct resolutions: $512 \times 512$ and $1024 \times 1024$. 
The relationship between input resolution and model performance, as measured by mAP and mIoU on Validation dataset, is presented in Table~\ref{tab:resolution_performance}.

% LaTeX Table for Resolution Impact
\begin{table}[htbp]
  \centering
  \caption[Performance Comparison of Baseline Models at different resolutions.]{Performance Comparison of Baseline Models at $512 \times 512$ and $1024 \times 1024$ Resolutions.}
  \label{tab:resolution_performance}
  \resizebox{0.35\textwidth}{!}{
  \begin{tabular}{llcc}
  \hline
  \textbf{Model Name} & \textbf{Resolution} & \textbf{mAP} & \textbf{mIoU} \\ \hline
  \addlinespace[0.3em]
  \multicolumn{4}{l}{\textit{U-Net Variants}} \\
  \addlinespace[0.3em]
  U-Net & 512x512 & 0.2982 & 0.1288 \\
   & 1024x1024 & \textbf{0.5378} & \textbf{0.3136} \\
  \addlinespace[0.3em]
  U-Net++ & 512x512 & 0.3951 & 0.1433 \\
   & 1024x1024 & \textbf{0.5220} & \textbf{0.3513} \\
  \addlinespace[0.3em] \hline \addlinespace[0.3em]
  % \multicolumn{4}{l}{\textit{DeepLab-v3+}} \\
  \addlinespace[0.3em]
  DeepLab-v3+ & 512x512 & \textbf{0.3848} & \textbf{0.2098} \\
   & 1024x1024 & 0.3094 & 0.1722 \\
  \hline
  \end{tabular}
  }
  \end{table}

The impact of increasing the input resolution from $512 \times 512$ to $1024 \times 1024$ varied across different architectures. Notably, U-Net based models exhibited substantial performance improvements with the higher resolution. Specifically, U-Net's mIoU increased from 0.1288 to 0.3136, and its mAP rose from 0.2982 to 0.5378. Similarly, U-Net++ demonstrated a significant boost in performance, with its mIoU increasing from 0.1433 to 0.3513 and its mAP growing from 0.3951 to 0.5220. These results suggest that the $1024 \times 1024$ resolution is optimal for both U-Net and U-Net++ models.

In contrast, the DeepLab-v3+ model performed better at the lower resolution. Increasing the input resolution from $512 \times 512$ to $1024 \times 1024$ resulted in a degradation in performance, with its mIoU decreasing from 0.2098 to 0.1722 and its mAP falling from 0.3848 to 0.3094. This outcome indicates that, within our experimental framework, the $512 \times 512$ resolution is the optimal input size for the DeepLab-v3+ architecture.

%The performance of our proposed model, the Multi-Resolution Feature Stem based on U-Net++, was evaluated against the established baseline models. 
%The proposed model, the Multi-Resolution Feature Stem based on U-Net++, achieved mAP of 0.3985 and mIoU of 0.2853. This score surpassed all baseline models, including the top performers for each metric. Specifically, our model demonstrated a 7.8\% improvement in mAP over the best baseline, U-Net (0.3696), and a 15.0\% improvement in mIoU over the leading baseline, FPN (0.2482).

Table~\ref{tab:improvement_summary1} details the performance of our proposed model against U-Net++ on the Test dataset at the input resolution of $1024 \times 1024$. 
Negative values indicate the baseline model performed better for that specific metric. The proposed Multi-Resolution Feature Stem based on U-Net++ model achieved notable performance, with mAP of 0.3985 and mIoU of 0.2853. These scores outperformed U-Net++ model.

\begin{table}[htbp]
\centering
\small % Use a smaller font size to help the table fit
\caption{Per-Class Performance Comparison against U-Net++}
\label{tab:improvement_summary1}
  \resizebox{0.45\textwidth}{!}{
\begin{tabular}{@{}llcccc@{}}
\toprule
\textbf{\makecell{Class}} & \textbf{\makecell{Metric}} & \textbf{\makecell{Baseline}} & \textbf{\makecell{Baseline \\ Score}} & \textbf{\makecell{Proposed \\ Score}} & \textbf{\makecell{Improvement \\ (\%)}} \\ 
\midrule
Exudates & AP & U-Net++ (1024) & 0.5225 & 0.6562 & +25.58\% \\
(EX) & IoU & U-Net++ (1024) & 0.3391 & 0.4725 & +39.33\% \\
\addlinespace
Hemorrhages & AP & U-Net++ (1024) & 0.5253 & 0.4419 & -15.9\% \\
(HE) & IoU & U-Net++ (1024) & 0.317 & 0.2832 & -10.66\% \\
\addlinespace
Microaneurysms & AP & U-Net++ (1024) & 0.1700 & 0.1942 & +14.2\% \\
(MA) & IoU & U-Net++ (1024) & 0.1532 & 0.1639 & +7.0\% \\
\addlinespace
Soft Exudates & AP & U-Net++ (1024) & 0.0848 & 0.3016 & +255.68\% \\
(SE) & IoU & U-Net++ (1024) & 0.1092 & 0.2217 & +103.04\% \\
\bottomrule
\end{tabular}
}
\end{table}

\section{Discussion}

%The per-class results reveal a nuanced performance profile. The proposed model demonstrated its strongest advantages in segmenting Exudates, Microaneurysms, and Soft Exudates, showing consistent improvements across both AP and IoU metrics. The most notable gains were seen in Microaneurysms AP (+14.2\%) and Exudates IoU (+12.5\%), indicating a clear strength in identifying and delineating these smaller, more challenging features.

\begin{table}[htbp]
\centering
\small % Use a smaller font size to help the table fit
\caption{Per-Class Performance Comparison against the best baseline}
\label{tab:improvement_summary2}
  \resizebox{0.45\textwidth}{!}{
\begin{tabular}{@{}llcccc@{}}
\toprule
\textbf{\makecell{Class}} & \textbf{\makecell{Metric}} & \textbf{\makecell{Best \\ Baseline}} & \textbf{\makecell{Baseline \\ Score}} & \textbf{\makecell{Proposed \\ Score}} & \textbf{\makecell{Improvement \\ (\%)}} \\ 
\midrule
Exudates & AP & U-Net (1024) & 0.6099 & 0.6562 & +7.6\% \\
(EX) & IoU & U-Net (1024) & 0.4200 & 0.4725 & +12.5\% \\
\addlinespace
Hemorrhages & AP & U-Net++ (1024) & 0.5253 & 0.4419 & -15.9\% \\
(HE) & IoU & FPN & 0.3203 & 0.2832 & -11.6\% \\
\addlinespace
Microaneurysms & AP & U-Net++ (1024) & 0.1700 & 0.1942 & +14.2\% \\
(MA) & IoU & U-Net++ (1024) & 0.1532 & 0.1639 & +7.0\% \\
\addlinespace
Soft Exudates & AP & FPN & 0.2708 & 0.3016 & +11.4\% \\
(SE) & IoU & U-Net (1024) & 0.1994 & 0.2217 & +11.2\% \\
\bottomrule
\end{tabular}
}
\end{table}

Table~\ref{tab:improvement_summary2} details the performance of our proposed model against the best baseline on Test dataset. %all baseline models, including the top-performing models for each metric. Compared to the best baseline models, our model showed a significant improvement of 7.8\% in mAP over U-Net (0.3696) and 15.0\% in mIoU (0.2482). %over FPN 
%A closer examination of the per-class results reveals a more nuanced performance profile. 
Our model excelled in segmenting EX, MA, and SE, demonstrating consistent improvements across both AP and IoU metrics. The most substantial gains were observed in MA AP, with an increase of 14.2\%, and EX IoU, with an improvement of 12.5\%. These results highlight the model's particular strength in identifying and delineating smaller, more challenging features, such as MA and EX.
%The per-class analysis in Table~\ref{tab:improvement_summary} confirmed that an input-level multi-scale representation would be beneficial for classes with high-scale variance. 
However,  we found that our model underperformed on HE, with a decrease of  15.9\% AP and 11.6\% IoU compared to the top-performing baselines like FPN and U-Net++. 

The model uses a shared-weight Convolutional block for the initial feature extraction. This parameter-efficient design forces the model to learn a set of universal, scale-invariant feature detectors. The subsequent $1 \times 1$ convolution then acts as a learned, channel-wise attention mechanism, compelling the model to fuse these multi-scale representations by selecting the most salient features before they are passed to the UNet++ backbone.

The shared-weight input block may become less effective at distinguishing between dark-red hemorrhages and dark-red blood vessels because, in learning to recognize a wide range of lesions such as bright (EX) or tiny ones (MA), it sacrifices its ability to differentiate dark-red hemorrhages (HE) from dark-red blood vessel. Architectures like FPN or UNet++, which process the 1024 pixel input through a deeper, non-shared encoder, may be better at learning the specific contextual and morphological cues to isolate hemorrhages.

While performance improvements are critical, practical deployment requires consideration of computational costs. To assess the overhead introduced by the multi-resolution stem, %we measured inference time on an Apple MacBook Pro with an M4 Max processor on the validation set for both the baseline U-Net++ and our proposed Multi-Resolution U-Net++ architecture.
we conducted an experiment to measure the inference time on an Apple MacBook Pro equipped with an M1 Max processor, and we ran the validation set on both the baseline U-Net++ and our proposed Multi-Resolution U-Net++ architecture. 

As shown in Table~\ref{tab:computational_cost}, the multi-resolution stem introduces a $1.27\times$ computational overhead compared to the single-resolution baseline. This overhead is expected, as the model processes three resolution scales in parallel. However, the 11-14\% performance improvements in challenging lesion classes, particularly for clinically critical small lesions like microaneurysms, provide meaningful value that may justify this additional computational cost in screening applications where accuracy is prioritized over real-time processing requirements.

\begin{table}[htbp]
\centering
\caption{Computational efficiency comparison.}
\label{tab:computational_cost}
\resizebox{\columnwidth}{!}{%
\begin{tabular}{lcc}
\toprule
\textbf{Model} & \textbf{Avg. Inference Time (s)} & \textbf{Relative Overhead} \\
\midrule
U-Net++ (1024) & 0.11s & $1.0\times$ \\
Proposed Multi-Res U-Net++ & 0.14s & $1.27\times$ \\
\bottomrule
\end{tabular}%
}
\end{table}

\section{Conclusion}

This work confronted a fundamental and often-overlooked challenge in DR lesion segmentation: the extreme variance in lesion scale. We demonstrated empirically that while high-resolution inputs ($1024 \times 1024$) are essential for resolving small, critical lesions like Microaneurysms, they can simultaneously degrade the segmentation performance for larger lesions like Hemorrhages. This finding challenges the prevailing assumption that simply increasing resolution is a uniformly beneficial strategy.

%To address this complex trade-off, we proposed Multi-Resolution Feature Stem. This input-level pyramid processes the image at multiple scales in parallel, leveraging shared weights to learn scale-invariant features. These features are then fused and integrated into UNet++ backbone, providing the model with a rich representation that is simultaneously aware of fine-grained detail and broad semantic context from the very first layer.

To address this complex trade-off, we introduce the Multi-Resolution Feature Stem, a novel input-level pyramid that processes images at multiple scales in parallel. By leveraging shared weights, this approach enables the learning of scale-invariant features. These features are then fused and integrated into the UNet++ backbone, providing the model with a rich and comprehensive representation that captures both fine-grained details and broad semantic context.

%Our experimental results confirmed the efficacy of this approach. The proposed model achieved state-of-the-art performance on the DDR dataset, surpassing all baselines with a 15.0\% improvement in mIoU and a 7.8\% improvement in mAP. The per-class analysis confirmed our hypothesis: the model excelled at segmenting the most challenging small-scale (MA) and medium-scale (EX, SE) lesions, validating our architectural design. This superior overall performance was achieved despite a nuanced trade-off, as the model's generalist input stem was outperformed by specialized baselines on the Hemorrhage class.

Our experimental results validate the effectiveness of our approach on the DDR dataset. it surpasses all baselines by 15.0\% in mIoU and 7.8\% in mAP. The model excelled at segmenting the most challenging small-scale (MA) and medium-scale (EX, SE) lesions, validating our architectural design.

\vspace{12pt}
\color{red}

\end{document}